\documentclass[journal]{IEEEtran}
\usepackage{cite}
\usepackage{amsmath,amssymb,amsfonts}
\usepackage{algorithmic}
\usepackage{graphicx}
\usepackage{textcomp}
\usepackage{booktabs}
\usepackage{tabularx}
\usepackage{multirow}
\usepackage{pifont}
\usepackage[hidelinks]{hyperref} 
\newcommand{\cmark}{\ding{51}}

\def\BibTeX{{\rm B\kern-.05em{\sc i\kern-.025em b}\kern-.08em
    T\kern-.1667em\lower.7ex\hbox{E}\kern-.125emX}}

\begin{document}
\title{CurConMix+: A Unified Spatio-Temporal Framework for Hierarchical Surgical Workflow Understanding}

\author{Yongjun Jeon\textsuperscript{1,2,*}, Jongmin Shin\textsuperscript{2,3,*}, Kanggil Park\textsuperscript{2,3,*}, Seonmin Park\textsuperscript{2,3}, Soyoung Lim\textsuperscript{3}, Jung Yong Kim\textsuperscript{3},\\
Jinsoo Rhu\textsuperscript{3}, Jongman Kim\textsuperscript{3}, Gyu-Seong Choi\textsuperscript{3}, Namkee Oh\textsuperscript{2,3,$\dagger$}, Kyu-Hwan Jung\textsuperscript{1,2,$\dagger$}\\[1em]
\textsuperscript{1}Department of Medical Device Management and Research, Samsung Advanced Institute for Health\\
Sciences \& Technology (SAIHST), Sungkyunkwan University, Seoul, Republic of Korea\\
\textsuperscript{2}Clinical Robotics and Embodied AI Research Center, Smart Healthcare Research Institute, Research\\
Institute for Future Medicine, Samsung Medical Center, Seoul, Republic of Korea\\
\textsuperscript{3}Department of Surgery, Samsung Medical Center, Seoul, Republic of Korea\\
Seoul, Republic of Korea\\[0.5em]
\textsuperscript{*}Equal contribution. \textsuperscript{$\dagger$}Corresponding authors.
\thanks{This work was supported by a grant of the Korean ARPA-H Project through the Korea Health Industry Development Institute (KHIDI), funded by the Ministry of Health \& Welfare, Republic of Korea (RS-2025-25424639); by the National Research Foundation of Korea (NRF) grant funded by the Korean government (Ministry of Science and ICT) (RS-2024-00392495); by the Future Medicine 2030 Project of Samsung Medical Center (SMX1230771); and by a grant from Samsung Medical Center (SMO1250271).}%
}

\maketitle
\begin{abstract}
Surgical action triplet recognition aims to understand fine-grained surgical behaviors by modeling the interactions among instruments, actions, and anatomical targets. Despite its clinical importance for workflow analysis and skill assessment, progress has been hindered by severe class imbalance, subtle visual variations, and the semantic interdependence among triplet components. Existing approaches often address only a subset of these challenges rather than tackling them jointly, which limits their ability to form a holistic understanding. This study builds upon \textbf{CurConMix}, a spatial representation framework. At its core, a curriculum-guided contrastive learning strategy learns discriminative and progressively correlated features, further enhanced by structured hard-pair sampling and feature-level mixup. Its temporal extension, \textbf{CurConMix+}, integrates a Multi-Resolution Temporal Transformer (MRTT) that achieves robust, context-aware understanding by adaptively fusing multi-scale temporal features and dynamically balancing spatio-temporal cues. Furthermore, we introduce \textbf{LLS48}, a new, hierarchically annotated benchmark for complex laparoscopic left lateral sectionectomy, providing step-, task-, and action-level annotations. Extensive experiments on \textbf{CholecT45} and \textbf{LLS48} demonstrate that \textbf{CurConMix+} not only outperforms state-of-the-art approaches in triplet recognition, but also exhibits strong cross-level generalization, as its fine-grained features effectively transfer to higher-level phase and step recognition tasks. Together, the framework and dataset provide a unified foundation for hierarchy-aware, reproducible, and interpretable surgical workflow understanding. The code and dataset will be publicly released on GitHub to facilitate reproducibility and further research.
\end{abstract}


\begin{figure*}
\centering
\includegraphics[width=\linewidth]{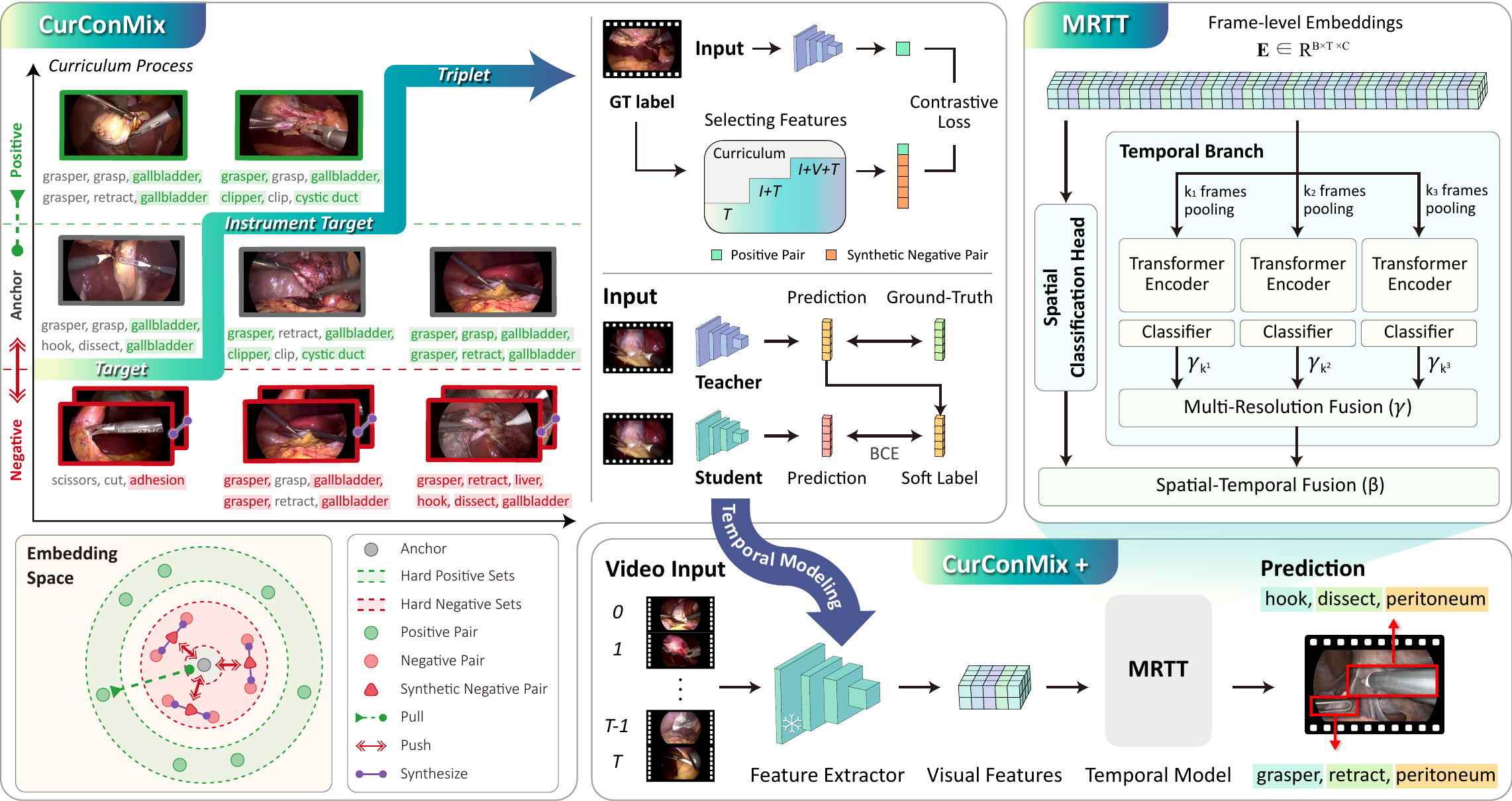}
\caption{\textbf{Overview of CurConMix+.}
The framework comprises two core modules: (1) the CurConMix spatial backbone and (2) the MRTT temporal module.
CurConMix employs curriculum-guided contrastive learning with hard-pair sampling and feature-level mixup to learn discriminative spatial features. 
The pretrained backbone is then fine-tuned via a self-distillation scheme incorporating input mixup to mitigate class imbalance.
Subsequently, the MRTT module captures temporal dynamics through parallel temporal pathways, whose outputs are adaptively fused using $\gamma_k$ and dynamically weighted with spatial cues via $\beta$.
The complete CurConMix+ pipeline integrates these spatial and temporal components to achieve robust spatio-temporal triplet prediction.}
\label{fig:fig2_curconmix_overview}
\end{figure*}

\section{Introduction}
Surgical decision-making has long relied on subjective expertise due to the lack of systematic intraoperative documentation~\cite{birkmeyer2013surgical, scally2016video, varban2021evaluating}.  
The advent of minimally invasive surgery (MIS) transformed this paradigm by enabling routine video-based recording~\cite{peter2008history}, turning surgery into a data-rich domain for quantitative process analysis.  
Early efforts in surgical AI focused on phase recognition, which provides a high-level understanding by segmenting procedures into major stages~\cite{padoy2012statistical,tecno,Trans_SVNet,twinanda2016endonet,SV-RCNet}.
Building on this, recent studies have advanced to fine-grained modeling, such as surgical action triplet recognition~\cite{nwoye2020recognition}, to capture more detailed interactions.  
This approach explicitly represents interactions as $\langle \text{instrument, verb, target} \rangle$, offering a structured representation that captures compositional dependencies for a more precise and interpretable understanding of intraoperative behavior.

Surgical action triplet recognition poses several inherent challenges.
First, a substantial class imbalance exists within surgical datasets, where a few frequent triplet combinations account for nearly half of all instances, causing head classes to dominate learning and hindering balanced representation learning.
Second, subtle intra- and inter-class variations make discrimination difficult, as visually similar scenes or overlapping tool–tissue interactions often exhibit minimal appearance differences.
Third, capturing the complex interdependencies among triplet components remains challenging.

Although existing approaches have achieved remarkable progress, they still struggle to tackle these challenges in a unified and coherent manner. 
Knowledge distillation methods such as SelfD~\cite{selfd} and TOD-Large~\cite{yamlahi2025smarter} alleviate class imbalance by reducing overconfidence in dominant classes. 
However, they neither capture subtle visual variations nor explicitly model dependencies among components. 
Meanwhile, multi-task learning (MTL) frameworks like TERL~\cite{gui2024tail} incorporate contrastive objectives to handle both imbalance and fine-grained differences. 
Yet, their use of separate prediction branches for each component in a triplet makes it difficult to model relationships among components, which are essential for compositional reasoning. 
Graph-based models~\cite{zou2025capturing}, in turn, explicitly encode component dependencies using dataset priors, but these priors are inherently biased by label imbalance and fail to learn discriminative features for subtle variations. 
Collectively, these limitations reveal that most existing methods address only specific aspects of the problem, suggesting the need for a unified framework that comprehensively tackles the core challenges.

Building on these observations, we first introduce CurConMix—a \textbf{Cur}riculum-based \textbf{Con}trastive learning framework with feature \textbf{Mix}up—as a unified framework designed to address all three challenges.
The core of the framework is curriculum-guided contrastive learning.
Our approach learns highly discriminative representations to separate visually similar triplets.
It also progressively models the complex semantic relationships between components, enhanced by progressive hard-pair sampling and feature-level mixup.
To specifically address the severe class imbalance and annotation ambiguity, the CurConMix framework then employs self-distillation with soft labels and input mixup during the fine-tuning stage.
Finally, to ensure temporally coherent and discriminative predictions, we incorporate a Multi-Resolution Temporal Transformer (MRTT) on top of the learned embeddings. This complete, unified framework, referred to as \textbf{CurConMix+}, is illustrated in Fig.~\ref{fig:fig2_curconmix_overview}.

In addition, we introduce \textbf{LLS48}, a new benchmark comprising 48 laparoscopic left lateral sectionectomy procedures. 
By annotating each procedure at three distinct levels—step, task, and action—LLS48 enables a comprehensive analysis of the surgical workflow across its hierarchical levels, and complements existing cholecystectomy benchmarks with a more complex hepatic procedure.

We evaluate \textbf{CurConMix+} on both the public CholecT45 dataset~\cite{cholecT45Dataset} and the newly introduced LLS48 dataset. 
The framework achieves 43.4\% AP\textsubscript{\textit{IVT}} on CholecT45 and 59.0\% on LLS48, outperforming existing methods across both benchmarks. 
Furthermore, the learned representations transfer effectively from fine-grained triplet recognition to higher-level workflow tasks—including phase, step, and task recognition—achieving 93.3\% phase accuracy on Cholec80 and 85.8\% step accuracy on LLS48, surpassing baselines and thereby confirming robust generalization across both spatial and temporal dimensions of surgical understanding.

In summary, the contributions of this work are as follows:
\begin{enumerate}
    \item We propose \textbf{CurConMix+}, a unified framework built upon CurConMix~\cite{jeon2025curconmix}, which extends frame-level representations by introducing a \textbf{Multi-Resolution Temporal Transformer (MRTT)} to learn temporally-coherent action representations.
    \item We introduce \textbf{LLS48}, a new benchmark for laparoscopic left lateral sectionectomy featuring 48 procedures with hierarchical annotations at three levels (\textit{step}, \textit{task}, \textit{action}), enabling standardized cross-level evaluation.
    \item We demonstrate through comprehensive experiments that \textbf{CurConMix+} achieves state-of-the-art performance on frame-level surgical action triplet recognition on both \textbf{LLS48} and the public \textbf{CholecT45} dataset, and that its learned action-level representations robustly transfer to higher-level hierarchical workflow recognition tasks.
\end{enumerate}

\section{Related Work}\label{sec:related}
\subsection{Surgical Action Triplet Recognition}
Surgical action triplet recognition aims to identify fine-grained surgical activities within video frames, represented as structured triplets of $\langle \text{instrument, verb, target} \rangle$. This task is a critical prerequisite for developing computer-assisted surgery systems and intelligent surgical environments, as it provides real-time situational awareness and supports surgical documentation and training, ultimately enhancing patient safety and efficiency. However, this task presents unique challenges: (1) a substantial class imbalance, (2) difficulty in discriminating subtle visual variations, and (3) the complexity of modeling interdependencies among triplet components. Existing research has attempted to solve these problems primarily through multi-task learning (MTL), knowledge distillation, and graph-based approaches.

Multi-task learning is the most common paradigm in surgical action triplet recognition, typically co-optimizing the full triplet prediction with its constituent component tasks. RDV~\cite{nwoye2022rendezvous} jointly trained these four tasks, attempting to capture component relationships via attention, while RiT~\cite{sharma2023rendezvous} extended this with temporal modeling. Recent studies leverage the MTL framework to address class imbalance. TERL~\cite{gui2024tail} combines an MTL base with instance-level contrastive learning to improve tail class recognition. MT4-MTL~\cite{gui2023mt4mtl} also utilizes an MTL structure to transfer knowledge from less imbalanced sub-tasks (I, V, T) to the main triplet task (IVT). Similarly, LAM~\cite{li2024parameter} uses hierarchical MTL to predict static elements (I, T) before dynamic ones (V, IVT). However, these MTL approaches share a fundamental limitation that by decomposing triplet prediction into separate branches, they may improve component-level performance but struggle to model the contextual coupling and interdependencies essential for compositional reasoning.

Knowledge Distillation approaches have been effective, primarily focusing on addressing class imbalance and label ambiguity. SelfD~\cite{selfd} uses a teacher model's soft labels to train a student, mitigating overconfidence in head classes. Going further, MT4-MTL~\cite{gui2023mt4mtl} employs multi-teacher distillation, while TOD-Large~\cite{yamlahi2025smarter} adopts a complex strategy of ensembling teacher checkpoints across multiple seeds and epochs to generate optimal soft labels. This strategy, however, requires significant computational resources. Critically, while these distillation-centric methods contribute to mitigating imbalance, they do not directly address the challenges of learning highly discriminative representations to resolve subtle visual differences or of explicitly modeling component dependencies.

Graph-based methods~\cite{xi2022forest,zou2025capturing} attempt to explicitly model the structural dependencies between triplet components to enhance features. However, these approaches often rely on dataset-level co-occurrence priors, which themselves are inherently biased by the original data's severe class imbalance, limiting their generalization. Meanwhile, multi-modal approaches like MEJO~\cite{zhang2025mejo} try to resolve semantic ambiguity by integrating expert knowledge generated from MLLMs. While this improves performance, it creates a dependency on external, large-scale models and requires a complex pipeline.

These existing studies, therefore, tend to focus on specific aspects of the three core challenges. This work introduces \textbf{CurConMix} and its temporal extension \textbf{CurConMix+} as a unified framework that cohesively addresses these challenges.

\section{Methods}\label{sec:methods}
\subsection{Overview of CurConMix}
The pretraining stage is designed to learn discriminative and semantically rich feature representations specialized for surgical action triplets. The framework integrates three complementary components: (\textit{i}) curriculum contrastive learning, (\textit{ii}) progressive hard pair sampling, and (\textit{iii}) synthetic hard negative generation.

\subsubsection{Curriculum Contrastive Learning}
Each sample is represented by a feature vector $f_i \in \mathbb{R}^d$ with a multi-label vector $\mathbf{y}_i \in \{0,1\}^C$ over $C$ triplet classes. Surgical triplet prediction is challenging due to strong semantic coupling among $\langle \text{instrument, verb, target} \rangle$ and the long-tailed, multi-label distribution. To address this, we adopt curriculum-guided supervised contrastive learning that progressively increases semantic granularity.

Conceptually, the curriculum refines the embedding space from coarse to fine constraints. Early stages use coarse equivalence (e.g., shared targets) to encourage global alignment and reduce within-group dispersion, yielding a more coherent structure. Later stages condition positives on instruments and verbs, refining coarse clusters into subclusters and deferring complex boundaries until the global organization stabilizes. This coarse-to-fine schedule helps avoid premature fragmentation in long-tailed multi-label settings and improves separability and clustering stability at finer semantics.

Following the surgical logic that a verb is defined by the interaction between the instrument and target~\cite{Cholectriplet2021}, we define three stages:
$\mathcal{S}_1\!:\!\langle \text{target} \rangle 
\rightarrow \mathcal{S}_2\!:\!\langle \text{instrument, target} \rangle 
\rightarrow \mathcal{S}_3\!:\!\langle \text{instrument, verb, target} \rangle.$
At stage $\mathcal{S}_k$, supervision is restricted to the label subset $\mathcal{Y}_{\mathcal{S}_k} \subset \mathcal{Y}$. The encoder $\mathcal{F}_\theta$ is optimized with the supervised contrastive loss~\cite{Supcon}:
\begin{equation}
\mathcal{L}_{\mathcal{S}_k} = 
\sum_{i \in I} -\frac{1}{|\mathcal{P}(i)|}
\sum_{p \in \mathcal{P}(i)}
\log
\frac{
\exp\!\left( z_i^\top z_p / \tau \right)
}{
\sum_{a \in \mathcal{A}(i)} 
\exp\!\left( z_i^\top z_a / \tau \right)
},
\label{eq:curriculum_supcon}
\end{equation}
where $z_i = f_i / \|f_i\|$, $\tau$ is a temperature, $\mathcal{P}(i)$ denotes positives sharing identical labels within $\mathcal{S}_k$, and $\mathcal{A}(i)$ includes all anchors in the batch. Overall, the stage-wise curriculum structures supervision from coarse to fine, enabling the encoder to capture component interdependencies while keeping the global embedding geometry stable.

\subsubsection{Progressive Pair Selection in Multi-Label Curriculum Contrastive Learning}
In contrastive learning, the selection of training pairs $(i,j)$ critically determines optimization dynamics~\cite{kalantidis2020hard}. 
Instead of random sampling or memory-bank augmentation~\cite{he2020momentum, chen2020simple}, 
we propose a hard pair sampling strategy that adapts to each curriculum stage $\mathcal{S}_k$ under multi-label supervision.

Let $s(i,j)$ denote the cosine similarity between two feature vectors:
\begin{equation}
s(i,j) = f_i^\top f_j / (\|f_i\| \, \|f_j\|).
\end{equation}
Given a stage $\mathcal{S}_k$, we define candidate positive and negative sets:
\begin{align}
\mathcal{H}_{\text{pos}}(i) &= \{ j \mid \mathbf{y}_{i,\mathcal{S}_k} = \mathbf{y}_{j,\mathcal{S}_k} \}, \\
\mathcal{H}_{\text{neg}}(i) &= \{ j \mid \mathbf{y}_{i,\mathcal{S}_k} \neq \mathbf{y}_{j,\mathcal{S}_k} \}.
\end{align}
From these, we identify hard candidates via similarity ranking:
\begin{align}
\mathcal{T}_{\text{pos}}^{K}(i) &= \operatorname*{BottomK}_{j \in \mathcal{H}_{\text{pos}}(i)} s(i,j), \\
\mathcal{T}_{\text{neg}}^{N}(i) &= \operatorname*{TopN}_{j \in \mathcal{H}_{\text{neg}}(i)} s(i,j).
\end{align}
We then sample uniformly within each hard pool to construct contrastive pairs:
\begin{align}
p_i &\sim \mathrm{Unif}\big(\mathcal{T}_{\text{pos}}^{K}(i)\big), \\
\mathcal{R}_{\text{neg}}(i) &\subset \mathcal{T}_{\text{neg}}^{N}(i), \quad |\mathcal{R}_{\text{neg}}(i)| = M.
\end{align}
Here, $K$, $N$, and $M$ are fixed caps (see Implementation Details). 
Sampling from $\mathcal{T}_{\text{pos}}^{K}(i)$ and $\mathcal{T}_{\text{neg}}^{N}(i)$ thus yields balanced yet challenging intra-class and inter-class pairs, 
focusing supervision on semantically ambiguous examples where visual similarity conflicts with label dissimilarity. 
As the curriculum progresses from $\mathcal{S}_1$ (e.g., distinguishing $\langle \text{target} \rangle$) to $\mathcal{S}_3$ (distinguishing full triplets), the criteria for defining positive pairs become progressively more stringent. 
Our hard pair sampling strategy adapts accordingly, forcing the model to focus on increasingly subtle semantic distinctions at each subsequent stage.

\subsubsection{Synthetic Hard Negative Feature Generation}
Although hard pair sampling enhances discriminative learning, the number of available hard negatives remains limited in long-tailed datasets. 
To further enrich the contrastive landscape, we introduce a synthetic hard negative generation mechanism using feature-level mixup.

For each anchor $i$, we randomly select two hard negative embeddings $v_{n_1}, v_{n_2} \in \mathcal{R}_{\text{neg}}(i)$ and interpolate them as:
\begin{equation}
\tilde{v}_s = \lambda v_{n_1} + (1-\lambda) v_{n_2},
\quad
\lambda \sim \text{Beta}(\alpha, \alpha),
\label{eq:synthetic_negative}
\end{equation}
where $\mathcal{R}_{\text{neg}}(i)$ is the uniformly sampled hard-negative subset defined above. 
This process constructs virtual negatives lying on the manifold between existing hard samples, 
introducing additional semantic diversity without requiring extra data.

Given one hard positive $p_i$ and $S$ synthetic negatives $\{\tilde{v}_s\}_{s=1}^S$ for anchor $i$, 
the supervised contrastive loss optimized at curriculum stage $k$ is formulated as:
\begin{equation}
\mathcal{L}_{\mathcal{S}_k} =
-\log
\frac{\exp (z_i^\top z_{p_i} / \tau)}
{\exp (z_i^\top z_{p_i} / \tau) +
\sum_{s=1}^{S} \exp (z_i^\top \tilde{v}_s / \tau)} ,
\label{eq:mix_loss_stage}
\end{equation}
where $z_i$ and $z_{p_i}$ are the normalized feature representations of the anchor and its hard positive, respectively.
The same contrastive objective is applied across all curriculum stages, with stage-specific granularity determining the pair definitions. 
By incorporating synthetic hard negatives into the contrastive space, 
the model progressively refines its discriminative boundaries and enhances robustness against unseen surgical variations.

\subsection{Mitigating Class Imbalance via Self-Distillation and Input Mixup}
Following pretraining, the model is fine-tuned for multi-label classification of surgical action triplets, addressing class imbalance and annotation ambiguity prevalent in surgical datasets such as CholecT45~\cite{cholecT45Dataset}. 
To alleviate overconfidence and noisy supervision, we adopt a self-distillation scheme~\cite{selfd} in which a teacher model $\mathcal{M}_T$ is first trained on hard labels using Binary Cross-Entropy (BCE), and a student model $\mathcal{M}_S$ of identical architecture learns from the teacher's soft predictions:
\begin{align}
\mathcal{L}_{\text{teacher}} 
&= -\frac{1}{C} \sum_{c=1}^{C} 
\big[
y_c \log(\hat{y}_c^{(T)}) 
+ (1 - y_c) \log(1 - \hat{y}_c^{(T)})
\big], \\
\mathcal{L}_{\text{student}} 
&= -\frac{1}{C} \sum_{c=1}^{C} 
\big[
\hat{y}_c^{(T)} \log(\hat{y}_c^{(S)}) 
+ (1 - \hat{y}_c^{(T)}) \log(1 - \hat{y}_c^{(S)})
\big],
\end{align}
where $\hat{y}_c^{(T)} = \mathcal{M}_T(x)_c$ and $\hat{y}_c^{(S)} = \mathcal{M}_S(x)_c$. 
Soft-label supervision transfers smoother, class-aware knowledge, improving calibration and generalization under long-tailed distributions~\cite{kim2021self, yang2023knowledge}. 

To further regularize fine-tuning, input-level mixup~\cite{mixup} is applied to both teacher and student. 
For two samples $(x_i, y_i)$ and $(x_j, y_j)$, mixed examples are generated as:
\begin{align}
\tilde{x} &= \lambda x_i + (1-\lambda) x_j, \\
\tilde{y} &= \lambda y_i + (1-\lambda) y_j,
\end{align}
where $\lambda \sim \text{Beta}(\alpha, \alpha)$ controls interpolation strength. 
The teacher trains on mixed hard labels $\tilde{y}$, while the student trains on mixed soft labels generated by the teacher, enforcing local linearity and smoothing decision boundaries.
Together, self-distillation and mixup stabilize optimization and enhance robustness to class imbalance and annotation noise, yielding more reliable and generalizable representations across surgical contexts.

\subsection{Multi-Resolution Temporal Modeling}
\label{sec:temporal_model}
Surgical workflows are temporally heterogeneous: fine-grained tool--tissue interactions unfold over short spans, whereas procedural intent evolves over longer horizons. To handle this, we propose a \textbf{Multi-Resolution Temporal Transformer (MRTT)} that models temporal reasoning through parallel pathways at multiple timescales. Each pathway models temporal patterns at a different timescale, and learnable fusion weights $\gamma_k$ assign sequence-level importance to these pathways, enabling the model to emphasize the motion patterns that best explain the procedural context. We use global (sequence-level) scale weights rather than frame-wise attention to reduce sensitivity to transient motion artifacts and to preserve procedural coherence.
While inspired by the temporal decoder in TOD-Large~\cite{yamlahi2025smarter}, MRTT replaces fixed averaging with learnable scale fusion and further incorporates trainable spatio-temporal weighting to balance appearance and motion cues within a unified framework.

Given frame-level embeddings $\mathbf{E} \in \mathbb{R}^{B \times T \times C}$ extracted from the student model's pretrained backbone, MRTT learns temporal dependencies through parallel pathways operating at different pooling strides.

\subsubsection{Temporal Pathways}
Each temporal pathway $\mathcal{P}_k$ applies average pooling with stride $k$ to reduce the temporal dimension while highlighting motion patterns at its respective resolution:
\begin{equation}
\mathbf{X}_k = \text{AvgPool}_{k}(\mathbf{E}), \quad
\mathbf{X}_k \in \mathbb{R}^{B \times T/k \times C}.
\end{equation}
The pooled features are encoded using an independent transformer encoder with fixed sinusoidal positional embeddings~\cite{vaswani2017attention},
which provide temporal ordering cues without introducing additional learnable parameters:
\begin{equation}
\mathbf{H}_k = \text{TransformerEnc}_k(\mathbf{X}_k + \text{PE}(\mathbf{X}_k)).
\end{equation}
The encoded features are then upsampled via temporal interpolation to restore the original sequence length:
\begin{equation}
\mathbf{H}_k^{\uparrow} = \text{Upsample}(\mathbf{H}_k, T).
\end{equation}
Each pathway outputs frame-wise logits through an independent classification head:
\begin{equation}
\mathbf{Z}_k = \mathbf{H}_k^{\uparrow} \mathbf{W}_{c,k} + \mathbf{b}_{c,k}.
\end{equation}
The pooling strides (e.g., $\{4,5,6\}$) provide diverse temporal receptive fields, enabling complementary motion representations at different temporal scales.

\subsubsection{Learnable Multi-Resolution Fusion}
Instead of uniform averaging, the model learns adaptive fusion weights $\gamma_k$ that modulate each temporal resolution's contribution:
\begin{equation}
\mathbf{Z}_{\text{temp}} = \sum_{k} \gamma_k \mathbf{Z}_k, \quad
\gamma_k = \frac{\exp(w_k)}{\sum_j \exp(w_j)}.
\end{equation}
Each $\gamma_k$ is a learnable scalar representing the global importance of the corresponding temporal resolution, allowing MRTT to emphasize the most relevant procedural patterns while suppressing less informative cues.

\subsubsection{Dynamic Spatio-Temporal Fusion}
In parallel, a spatial classification head predicts per-frame logits directly from the frame embeddings:
\begin{equation}
\mathbf{Z}_{\text{spat}} = \mathbf{E}\mathbf{W}_s + \mathbf{b}_s.
\end{equation}
The final prediction integrates both spatial and temporal evidence through a learnable convex combination:
\begin{equation}
\mathbf{Z}_{\text{final}} = \beta\,\mathbf{Z}_{\text{spat}} + (1-\beta)\,\mathbf{Z}_{\text{temp}}, \quad \beta \in [0,1],
\end{equation}
where $\beta$ is a global learnable scalar that adaptively balances spatial semantics and temporal dynamics, enhancing robustness to varying motion speeds and abrupt transitions.

\subsection{LLS48 Dataset}
\label{sec:LLS48}
Addressing the lack of a consistent hierarchical structure for Minimally Invasive Surgery (MIS) workflow analysis, recent efforts have sought to establish unified standards. The Society of American Gastrointestinal and Endoscopic Surgeons (SAGES) provided a foundational framework by introducing a four-level taxonomy comprising \textit{phase}, \textit{step}, \textit{task}, and \textit{action}~\cite{meireles2021sages}. 

Building directly on this standardized framework, we introduce \textbf{LLS48}, a hierarchical endoscopic video dataset for laparoscopic left lateral sectionectomy (LLS), designed to enable multi-level analysis of surgical workflows. 
LLS48 provides aligned annotations across three of these hierarchical levels—\textit{action}, \textit{task}, and \textit{step}.
The dataset comprises 48 complete LLS procedures performed by four hepatobiliary surgeons at Samsung Medical Center between February~2020 and March~2022.
All videos were anonymized, and the study was approved by the Institutional Review Board (IRB) of Samsung Medical Center (SMC-2023-04-128).

Videos were recorded at two resolutions (1920$\times$1080@60~FPS and 1356$\times$728@30~FPS), totaling approximately 52~hours of surgical footage. To ensure efficient yet representative sampling, 5-second clips were extracted every 90~seconds using \texttt{ffmpeg}~\cite{ffmpeg}, a rate empirically selected to balance temporal continuity and content diversity. After excluding out-of-body scenes, the final dataset includes 1,912 annotated clips (9,560 frames, 2~h~39~min~20~s).

Each clip is annotated at three hierarchical levels:
a \textit{step} denotes a procedure-specific stage with a clinical objective, a \textit{task} represents a reusable surgical subroutine applicable across procedures, and an \textit{action} corresponds to a triplet $\langle \text{instrument, verb, target} \rangle$ describing fine-grained tool–tissue interactions.
LLS48 contains 5~\textit{steps}, 18~\textit{tasks}, and 195~unique action triplets derived from 17~instruments, 16~verbs, and 38~targets. All levels exhibit a long-tailed distribution, reflecting the inherent imbalance and procedural diversity of real-world surgical workflows.

Annotation was performed by five experts—two hepatobiliary surgeons and three experienced scopists—with more than ten years of experience in minimally invasive surgery. 
Scopists conducted the initial labeling. Surgeons then reviewed and finalized these labels using a Streamlit-based dashboard designed to ensure consensus, consistency, and traceability.

\section{Experiments}
\label{sec:experiments}
\begin{table*}[t]
    \caption{Action Triplet Recognition Results of Different Methods on CholecT45 and LLS48}
  \label{tab:main_table}
  \centering
  \resizebox{\textwidth}{!}{
  \begin{tabular}{@{}l lccccccc@{}}
    \toprule
    Dataset & Method & Backbone & \( \text{AP}_I \) & \( \text{AP}_V \) & \( \text{AP}_T \) & \( \text{AP}_{IV} \) & \( \text{AP}_{IT} \) & \( \text{AP}_{IVT} \) \\
    \midrule
    \multirow{19}{*}{CholecT45} 
    & RDV \cite{nwoye2022rendezvous} & Res18          & 89.3$\pm$2.1 & 62.0$\pm$1.3 & 40.0$\pm$1.4 & 34.0$\pm$3.3 & 30.8$\pm$2.1 & 29.4$\pm$2.8 \\
    & RiT \cite{sharma2023rendezvous} & Res18+Attention & 88.6$\pm$2.6 & 64.0$\pm$2.5 & 43.4$\pm$1.4 & 38.3$\pm$3.5 & 36.9$\pm$1.0 & 29.7$\pm$2.6 \\
    & TDN \cite{chen2023surgical}     & Res50          & 91.2$\pm$1.9 & 65.3$\pm$2.8 & 43.7$\pm$1.6 & - & - & 33.8$\pm$2.5 \\
    & MT4MTL-KD \cite{gui2023mt4mtl}  & SwinL(384)+MS-TCT     & 93.1$\pm$2.1 & 71.8$\pm$3.4 & 48.8$\pm$3.8 & 44.9$\pm$2.4 & 43.1$\pm$2.0 & 37.1$\pm$0.5 \\
    & SelfD \cite{selfd}    & SwinB(224) $\dagger$ & 90.3$\pm$2.3 & 67.4$\pm$1.5 & 47.9$\pm$1.8 & 43.7$\pm$4.1 & 42.9$\pm$1.6 & 37.1$\pm$1.9 \\
    & SelfD \cite{selfd}    & SwinB$\times$2+SwinL & - & - & - & - & - & 38.5$\pm$0.0 \\
    & TERL-T \cite{gui2024tail}       & SwinT(224)+MSTCN $\dagger$ & 93.5$\pm$1.5 & 71.4$\pm$2.2 & 47.2$\pm$2.6 & 44.7$\pm$3.8 & 42.0$\pm$2.4 & 35.7$\pm$1.6 \\
    & TERL-B \cite{gui2024tail}       & SwinB(224)+MSTCN $\dagger$ & 93.9$\pm$2.0 & 70.8$\pm$2.3 & 49.4$\pm$4.7 & 43.9$\pm$3.4 & 43.6$\pm$2.6 & 35.6$\pm$1.4 \\
    & TERL-B \cite{gui2024tail}       & SwinB(384)+MSTCN $\dagger$ & 94.1$\pm$2.3 & 73.0$\pm$1.4 & 51.1$\pm$3.8 & 46.5$\pm$4.9 & 44.9$\pm$1.8 & 37.7$\pm$1.5 \\
    \cmidrule{2-9}
    & CurConMix-T \cite{jeon2025curconmix} & SwinT(224) $\dagger$          & 90.4$\pm$2.1 & 67.8$\pm$1.8 & 48.3$\pm$3.4 & 43.3$\pm$2.9 & 43.3$\pm$1.8 & 37.7$\pm$2.1 \\
    & CurConMix-B \cite{jeon2025curconmix} & SwinB(224) $\dagger$         & 90.4$\pm$3.0 & 68.2$\pm$1.5 & 49.7$\pm$2.5 & 44.8$\pm$5.4 & 45.3$\pm$2.4 & 38.8$\pm$2.8 \\
    & CurConMix-B \cite{jeon2025curconmix} & SwinB(384) $\dagger$         & 90.9$\pm$2.0 & 68.3$\pm$1.3 & 49.8$\pm$3.2 & 45.2$\pm$4.2 & 45.1$\pm$1.1 & 39.1$\pm$2.0 \\
    & CurConMix-L \cite{jeon2025curconmix} & SwinL(224) $\dagger$ & 91.4$\pm$2.0 & 68.5$\pm$1.0 & 48.9$\pm$2.2 & 44.5$\pm$3.4 & 44.9$\pm$1.6 & 39.1$\pm$1.9 \\
    \cmidrule{2-9}
    &  MEJO-B \cite{zhang2025mejo}       & SwinB+TransFPN & 93.9$\pm$2.2 & 72.9$\pm$1.8 & 51.6$\pm$5.7 & 47.8$\pm$6.4 & 46.3$\pm$1.8 & 41.2$\pm$2.6 \\
    &  TOD-Large \cite{yamlahi2025smarter} & SwinB+SwinL+Temporal & - & - & - & - & - & 41.4$\pm$0.0 \\
    & Zou et al. \cite{zou2025capturing}   & SwinB + Attention & 93.5$\pm$1.9 & 71.6$\pm$1.0 & 49.8$\pm$1.9 & 48.5$\pm$4.5 & 48.2$\pm$3.0 & 41.5$\pm$2.8 \\
    \cmidrule{2-9}
    & CurConMix-B+ & SwinB(224)+MRTT $\dagger$ & 93.6$\pm$1.8 & 70.5$\pm$1.6 & 50.5$\pm$3.2 & 45.9$\pm$3.6 & 48.7$\pm$2.3 & 41.6$\pm$2.8 \\
    & CurConMix-B+ & SwinB(384)+MRTT $\dagger$ & 93.4$\pm$2.3 & 73.1$\pm$0.7 & 51.5$\pm$4.1 & 47.6$\pm$4.1 & 47.6$\pm$2.4 & 41.7$\pm$2.4 \\
    & CurConMix-L+ & SwinL(224)+MRTT $\dagger$ & 93.3$\pm$1.4 & 72.1$\pm$1.7 & 49.7$\pm$1.7 & 47.7$\pm$3.5 & 48.5$\pm$0.4 & 42.2$\pm$1.8 \\
    & CurConMix-Ens+ & SwinB+SwinL+MRTT $\dagger$ & \textbf{94.1$\pm$1.9} & \textbf{73.9$\pm$1.3} & \textbf{52.0$\pm$3.0} & \textbf{49.2$\pm$4.3} & \textbf{49.6$\pm$1.5} & \textbf{43.4$\pm$2.2} \\
    \midrule
    \multirow{13}{*}{LLS48}
    & RDV & Res18 $\dagger$ & 48.6$\pm$6.9 & 46.8$\pm$7.9 & 36.6$\pm$5.4 & 30.7$\pm$3.5 & 29.5$\pm$4.2 & 27.7$\pm$4.4 \\
    & RiT & Res18 $\dagger$ & 42.9$\pm$3.5 & 42.1$\pm$4.4 & 34.7$\pm$2.3 & 29.3$\pm$2.8 & 28.1$\pm$1.7 & 27.9$\pm$1.8 \\
    & SelfD-B  & SwinB(224) $\dagger$ & 73.3$\pm$3.9 & 68.7$\pm$3.1 & 54.1$\pm$2.3 & 50.3$\pm$2.0 & 50.3$\pm$1.8 & 48.0$\pm$2.7 \\
    & TERL-T & SwinT(224)+MSTCN $\dagger$ & 72.1$\pm$1.8 & 65.8$\pm$0.7 & 52.0$\pm$2.4 & 47.4$\pm$4.1 & 46.3$\pm$1.1 & 43.3$\pm$1.5 \\
    & TERL-B & SwinB(224)+MSTCN $\dagger$ & 73.0$\pm$4.3 & 66.7$\pm$1.7 & 53.0$\pm$3.3 & 48.5$\pm$3.0 & 46.7$\pm$1.8 & 44.0$\pm$2.6 \\
    & TERL-B & SwinB(384)+MSTCN $\dagger$ & 76.4$\pm$3.4 & 68.5$\pm$2.0 & 55.1$\pm$3.7 & 51.2$\pm$4.1 & 49.6$\pm$1.3 & 45.9$\pm$2.5 \\
    \cmidrule{2-9}
    & CurConMix-T \cite{jeon2025curconmix} & SwinT(224) $\dagger$ & 71.7$\pm$4.1 & 66.9$\pm$3.4 & 54.6$\pm$2.5 & 47.9$\pm$2.7 & 49.4$\pm$1.8 & 46.9$\pm$2.0 \\
    & CurConMix-B \cite{jeon2025curconmix} & SwinB(224) $\dagger$ & 74.1$\pm$3.2 & 68.3$\pm$2.1 & 55.9$\pm$2.0 & 52.0$\pm$2.1 & 52.4$\pm$1.3 & 49.7$\pm$1.9 \\
    & CurConMix-B \cite{jeon2025curconmix} & SwinB(384) $\dagger$ & 74.4$\pm$2.6 & 69.3$\pm$2.8 & 57.8$\pm$1.7 & 52.0$\pm$2.4 & 53.7$\pm$2.3 & 51.0$\pm$2.5 \\
    & CurConMix-L \cite{jeon2025curconmix} & SwinL(224) $\dagger$ & 74.1$\pm$2.7 & 68.9$\pm$2.6 & 57.1$\pm$2.3 & 51.8$\pm$2.2 & 52.9$\pm$1.8 & 50.5$\pm$2.0 \\
    \cmidrule{2-9}
    & CurConMix-B+ & SwinB(224)+MRTT $\dagger$ & 80.5$\pm$3.9 & 74.8$\pm$3.3 & 59.7$\pm$3.4 & 57.2$\pm$1.4 & 57.6$\pm$1.6 & 56.6$\pm$3.0 \\
    & CurConMix-B+ & SwinB(384)+MRTT $\dagger$ & \textbf{80.7$\pm$3.3} & 73.9$\pm$4.6 & 60.3$\pm$2.0 & 58.7$\pm$2.2 & 60.0$\pm$0.8 & 58.4$\pm$2.0 \\
    & CurConMix-L+ & SwinL(224)+MRTT $\dagger$ & 77.6$\pm$3.6 & 73.8$\pm$3.4 & 60.8$\pm$3.1 & 57.1$\pm$2.2 & 58.1$\pm$1.1 & 56.3$\pm$2.0 \\
    & CurConMix-Ens+ & SwinB+SwinL+MRTT $\dagger$ & 79.5$\pm$3.8 & \textbf{75.4$\pm$3.5} & \textbf{62.8$\pm$3.8} & \textbf{59.5$\pm$2.8} & \textbf{60.7$\pm$1.6} & \textbf{59.0$\pm$2.8} \\
    \bottomrule
  \end{tabular}
   }
\end{table*} 

\subsection{Datasets and Evaluation}
We evaluate the \textbf{CurConMix} framework under two complementary settings: 
(1) fine-grained surgical action triplet recognition, and 
(2) hierarchical workflow understanding across \textit{phase, step, and task levels}. 
These settings jointly assess the model's ability to learn compositional, interpretable, and transferable surgical representations.

\subsubsection{Surgical Action Triplet Recognition}
This task is evaluated on two datasets: CholecT45~\cite{cholecT45Dataset} and LLS48.  
CholecT45, a curated subset of Cholec80, contains 45 laparoscopic cholecystectomy videos with 90{,}489 annotated frames and 100 unique triplets. 
Each frame is labeled with one or more triplets $\langle \text{instrument, verb, target} \rangle$, supporting multi-label classification. 
LLS48, described in Section~\ref{sec:LLS48}, comprises 48 left lateral sectionectomy procedures annotated at three hierarchical levels—\textit{action}, \textit{task}, and \textit{step}—including 195 distinct triplets derived from 17 instruments, 16 verbs, and 38 targets. 
All experiments employ 5-fold cross-validation following prior studies~\cite{nwoye2022rendezvous,sharma2023rendezvous,chen2023surgical,gui2023mt4mtl,selfd,gui2024tail}.  
Performance is measured using Average Precision (AP) across component- and composite-level metrics: 
AP\textsubscript{\textit{I}}, AP\textsubscript{\textit{V}}, AP\textsubscript{\textit{T}}, AP\textsubscript{\textit{IV}}, AP\textsubscript{\textit{IT}}, and the primary triplet-level metric AP\textsubscript{\textit{IVT}}.  
These jointly evaluate recognition accuracy for individual components and compositional consistency of predicted triplets.
\begin{figure}[!htbp]
\centering
\includegraphics[width=0.9\columnwidth, keepaspectratio]{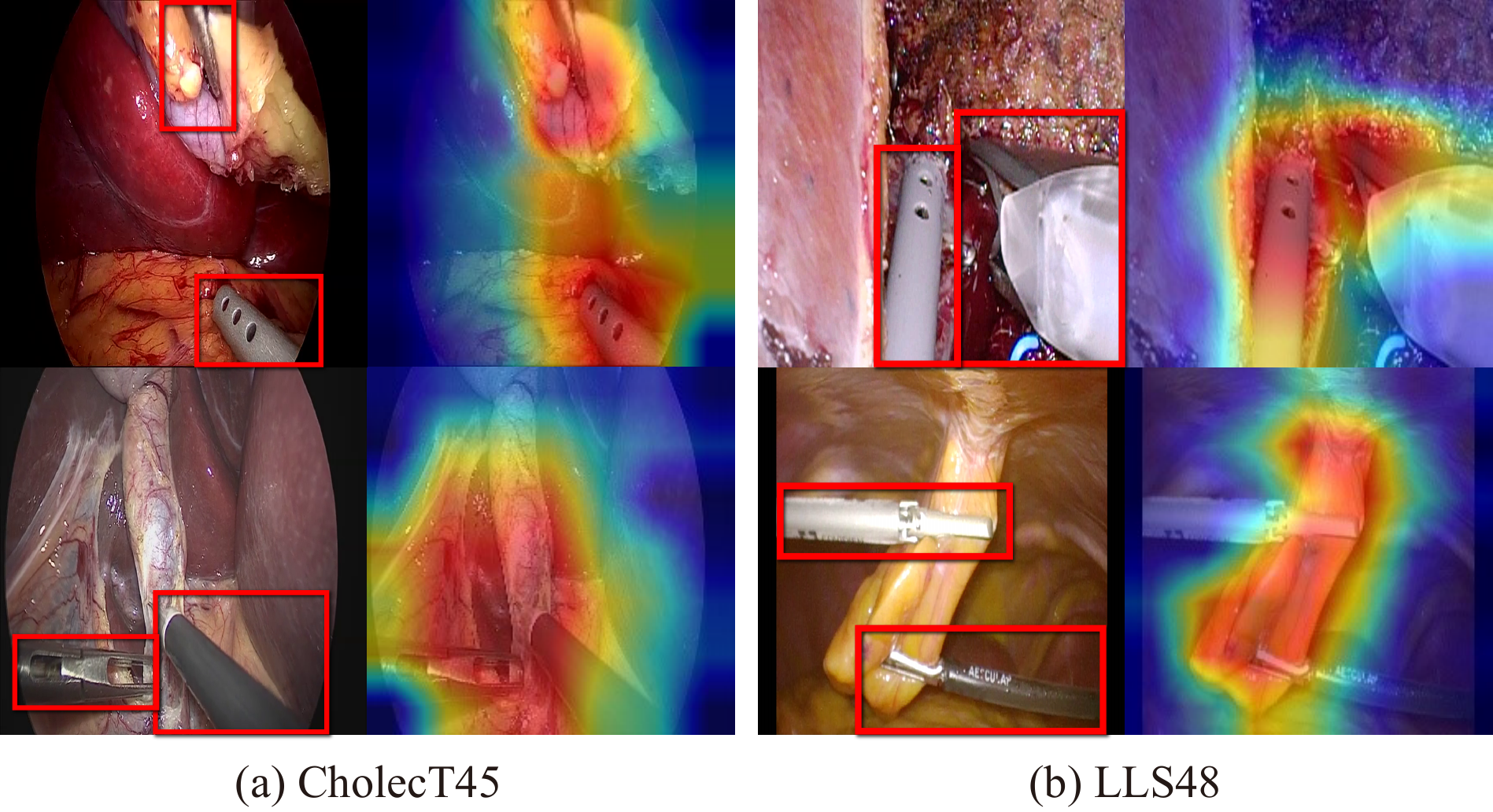}
\caption{\textbf{Grad-CAM visualization of CurConMix on CholecT45 and LLS48.}}
\label{fig:fig07_cam}
\end{figure}
\begin{figure}[t]
\centering
\includegraphics[width=\linewidth]{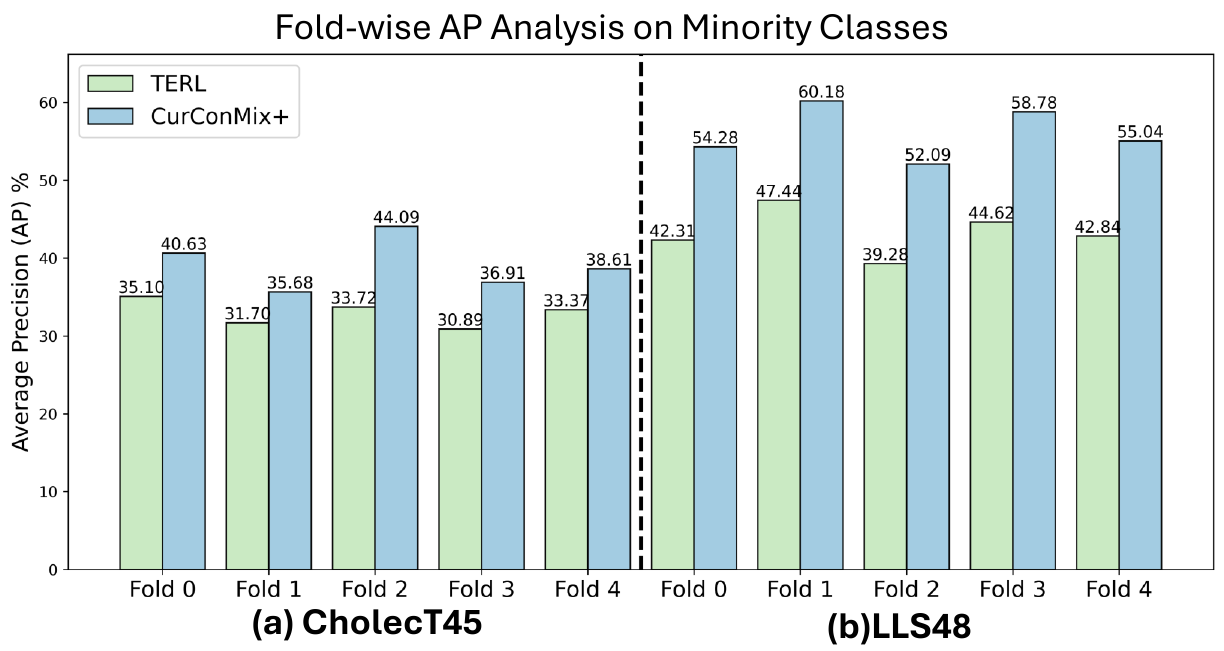}
\caption{\textbf{Comparison of minority-class performance between TERL and CurConMix+ on CholecT45 and LLS48.}}
\label{fig:fig_3_comp_TERL}
\end{figure}

\subsubsection{Hierarchical Workflow Understanding} 
To evaluate whether the fine-grained representations learned by \textbf{CurConMix} generalize to higher-level workflow understanding, we transfer the trained backbone to coarser-grained tasks involving \textit{phase}, \textit{step}, and \textit{task} recognition. The backbone remains frozen, and only the temporal model is trained for each of these higher-level tasks. Since the backbone was trained on CholecT45 (45 videos from Cholec80), we avoid data leakage by performing fine-tuning only on these 45 videos using a 5-fold cross-validation scheme, while the remaining 35 non-overlapping videos serve as a fixed test set. Five models are obtained from the cross-validation, and their performance on the 35 test videos is averaged to report the final results.
For step and task recognition, we additionally leverage the LLS48 dataset, which provides multi-level annotations (\textit{step}, \textit{task}, \textit{action}) and enables a direct assessment of cross-level generalization. We reuse the action-level splits to ensure a fair and consistent comparison across the annotation hierarchy.

Evaluation metrics include accuracy, precision, recall, and Jaccard index. For phase recognition, we additionally report the relaxed boundary metric from the MICCAI 2016 M2CAI challenge~\cite{twinanda2016endonet}. All splits are ensured to be non-overlapping between train and validation sets for unbiased benchmarking.

\subsection{Implementation Details}
All experiments are conducted on a single NVIDIA GeForce RTX~4090 GPU under identical hyperparameter settings for fair comparison across datasets and model variants.

For surgical action triplet recognition, all models employ Swin Transformer~\cite{liu2021swin} backbones (Tiny, Base, Large) pretrained on ImageNet~\cite{deng2009imagenet} using the \texttt{timm} library~\cite{rw2019timm}.  
Training uses AdamW with a learning rate of $2\times10^{-4}$ (decayed to $2\times10^{-5}$), weight decay of $1\times10^{-6}$, and batch size 64.  
Input- and feature-level mixup are applied with $\alpha=0.4$~\cite{mixup}, and the supervised contrastive temperature is fixed at $\tau=0.1$~\cite{Supcon}.  
CurConMix training proceeds in three stages: 3-epoch curriculum-based supervised contrastive pretraining, followed by teacher and student fine-tuning for 20 and 40 epochs, respectively, under the self-distillation scheme~\cite{selfd}.  
Temporal reasoning is integrated via the Multi-Resolution Temporal Transformer (Section~\ref{sec:temporal_model}), consisting of a 3-layer encoder with 4 attention heads, feedforward dimension 2048, dropout 0.1, and fixed sinusoidal positional encoding.  
Multi-scale temporal representations are constructed using masked average pooling with strides $k \in \{4,5,6\}$.  
Despite additional contrastive and temporal stages, the setup remains computationally efficient—adding less than 10\% training time over SelfD—while yielding stronger spatio-temporal representations.

For hierarchical workflow understanding, transfer learning follows a two-stage pipeline: (i) feature extraction using the triplet-pretrained CurConMix backbone (frozen during transfer), and (ii) temporal aggregation.  
Baseline Swin Transformer extractors pretrained on ImageNet are trained for Cholec80 and LLS48 step/task recognition under the TeCNO~\cite{tecno} multi-stage setup.  
Extracted features are then fed into temporal aggregation modules—TeCNO~\cite{tecno}, Trans-SVNet~\cite{Trans_SVNet}, and Not End-to-End~\cite{padoy2012statistical}—following their original configurations.

\section{Results}
\subsection{Results on Surgical Action Triplet Recognition in CholecT45 and LLS48}
Table~\ref{tab:main_table} summarizes the quantitative results on CholecT45 and LLS48, comparing the \textbf{CurConMix} and its temporal extension \textbf{CurConMix+} with recent state-of-the-art methods. 
Despite using a smaller backbone, CurConMix-T attains 37.7\% \( \text{AP}_{IVT} \) on CholecT45, outperforming MT4MTL-KD~\cite{gui2023mt4mtl}. 
Compared to SelfD~\cite{selfd}, CurConMix-B(224) achieves higher accuracy on both datasets, improving from 37.1\% to 38.8\% on CholecT45 and from 48.0\% to 49.7\% on LLS48. 
CurConMix also surpasses TERL~\cite{gui2024tail} across all backbones, yielding gains of up to +3.2 \( \text{AP}_{IVT} \) on CholecT45 and +5.7 on LLS48, even without explicit temporal reasoning. 
These consistent improvements indicate that the curriculum-based contrastive optimization in CurConMix enables the model to progressively capture interdependencies among instruments, verbs, and targets while producing highly discriminative spatial representations. 

When extended with temporal modeling, CurConMix+ further enhances performance by integrating temporal context. This improvement is clearly demonstrated on LLS48, where CurConMix-L+ achieves 56.3\% $\text{AP}_{IVT}$ and exceeds its non-temporal counterpart's 50.5\% by a significant +5.8 points.
Building on this, the full CurConMix+ model achieves 43.4\% $\text{AP}_{IVT}$ on CholecT45 and 59.0\% on LLS48. Consequently, CurConMix+ with 43.4\% surpasses all existing methods on CholecT45, including the 41.5\% from Zou~et~al.~\cite{zou2025capturing} and 41.4\% from TOD-Large~\cite{yamlahi2025smarter}. This strong performance stems from our unified design. The core CurConMix framework provides superior spatial representations that dynamically model compositional relationships, outperforming methods based on static priors~\cite{zou2025capturing}. This strong spatial foundation is also key to our advantage over TOD-Large~\cite{yamlahi2025smarter}. Whereas TOD-Large relies on a computationally heavy, multi-teacher ensemble for its knowledge distillation, our method achieves superior performance using a far more efficient, single-teacher distillation scheme. On LLS48, CurConMix-B+ attains 56.6\% $\text{AP}_{IVT}$, substantially improving upon TERL-B (44.0\%) by +12.6 points. As shown in Fig.~\ref{fig:fig_3_comp_TERL}, these gains are particularly pronounced for minority action triplets, where CurConMix+ consistently surpasses TERL across folds on both CholecT45 and LLS48, highlighting robustness to severe class imbalance and the long-tail distribution of rare triplets.

Qualitative visualizations further demonstrate that \textbf{CurConMix} achieves semantically grounded and clinically interpretable predictions.
As shown in Fig.~\ref{fig:fig07_cam}, Grad-CAM visualizations reveal that the model consistently attends to clinically relevant regions corresponding to annotated instruments, verbs, and targets. 
For instance, in CholecT45, it accurately localizes multiple concurrent actions within the same frame—such as $\langle$grasper, retract, gallbladder$\rangle$ and $\langle$irrigator, aspirate, fluid$\rangle$—while in LLS48, it distinguishes actions performed by different instruments on the same anatomical structure (e.g., round ligament). 
These spatially grounded attention maps highlight \textbf{CurConMix}'s ability to precisely disentangle complex surgical interactions—robustly discriminating distinct triplets across targets and consistently resolving concurrent actions on the same target—thereby achieving coherent and clinically meaningful visual reasoning.

\subsection{Comprehensive Ablation Analysis of CurConMix and CurConMix+}
We conducted ablation studies to assess the contribution of each component within the \textbf{CurConMix} framework and its temporal extension \textbf{CurConMix+}. 
All experiments were performed on the CholecT45 dataset using the official 5-fold validation splits with the SwinB(224) backbone under identical configurations. Table~\ref{tab:full_ablation_single} provides a unified summary of all ablation experiments. 
A clear stepwise improvement is observed as each module is added, demonstrating their complementary roles and cumulative impact on compositional reasoning.

\textbf{CurConMix Components.} 
Integrating supervised contrastive learning enhances feature discriminability, while curriculum learning promotes structured alignment among correlated components—\textit{instrument}, \textit{verb}, and \textit{target}—to strengthen compositional understanding.  
Input-level mixup improves robustness to label noise and mitigates class imbalance, whereas feature-level mixup—combining hard-pair sampling with synthetic hard negatives—achieves the best performance (38.8\%), demonstrating the benefit of training with harder and more diverse examples.  
Collectively, these components act complementarily, and the full configuration yields the highest accuracy when all modules are jointly optimized.

\textbf{CurConMix+ Components.} 
The lower section of Table~\ref{tab:full_ablation_single} further evaluates the temporal extension \textbf{CurConMix+}, which incorporates a Multi-Resolution Temporal Transformer with learnable fusion mechanisms. 
The effects of multi-resolution weighting (\(\gamma_k\)) and spatio-temporal balancing (\(\beta\)) are examined individually and jointly, both contributing to consistent performance gains. 
Adaptive resolution weighting through \(\gamma_k\) improves temporal reasoning by learning to emphasize relevant motion scales, while \(\beta\) dynamically balances spatial and temporal cues. 
When jointly optimized, the model achieves the highest mean AP\textsubscript{\textit{IVT}} of 41.04\%, indicating that learnable multi-resolution fusion effectively captures both short-term transitions and long-range procedural dependencies.

Overall, these ablation results confirm that both spatial–semantic and temporal modules are essential for comprehensive surgical scene understanding. 
While the base \textbf{CurConMix} framework establishes strong compositional representations through structured contrastive learning and mixup-based regularization, its temporal extension \textbf{CurConMix+} further enhances temporal coherence and stability by adaptively modeling multi-scale surgical dynamics.
\begin{table}[!t]
\centering
\renewcommand{\arraystretch}{1.15}
\caption{Ablation of CurConMix and CurConMix+ components on CholecT45}
\label{tab:full_ablation_single}
\resizebox{\columnwidth}{!}{%
{\small
\begin{tabular}{c c c c | c c c}
\toprule
\multicolumn{4}{c}{\textbf{CurConMix}} & \multicolumn{2}{c}{\textbf{CurConMix+}} & \\
\cmidrule(lr){1-4}\cmidrule(lr){5-6}
\textbf{SupCon} & \textbf{Curriculum} & \textbf{Input Mixup} & \textbf{Feature Mixup} & \textbf{Multi-Res (\(\gamma_k\))} & \textbf{Spatio-Temporal (\(\beta\))} & \textbf{AP\textsubscript{\textit{IVT}}} \\
\midrule
 & & & & & & 37.08$\pm$1.9 \\
\cmark & & & & & & 37.76$\pm$2.4 \\
\cmark & \cmark & & & & & 38.08$\pm$2.1 \\
\cmark & \cmark & \cmark & & & & 38.25$\pm$3.8 \\
\cmark & \cmark & \cmark & \cmark & & & 38.75$\pm$2.8 \\
\cmark & \cmark & \cmark & \cmark & \cmark & & 40.84$\pm$2.7 \\
\cmark & \cmark & \cmark & \cmark & & \cmark & 40.96$\pm$2.6 \\
\cmark & \cmark & \cmark & \cmark & \cmark & \cmark & \textbf{41.04$\pm$2.7} \\
\bottomrule
\end{tabular}
}}
\end{table}
\begin{table}[!t]
\centering
\small
\renewcommand{\arraystretch}{1.15}
\caption{
Effect of curriculum stage order on triplet recognition}
\label{tab:curriculum_order_mean_only}
\resizebox{\columnwidth}{!}{%
\begin{tabular}{ccc cc}
\toprule
\textbf{Stage 1} & \textbf{Stage 2} & \textbf{Stage 3} &
\textbf{CholecT45 (\(AP_{IVT}\))} & \textbf{LLS48 (\(AP_{IVT}\))} \\
\midrule
I & IV & IVT & 38.5$\pm$2.2 & 49.1$\pm$1.5\\
I & IT & IVT & 38.6$\pm$2.5 & 48.9$\pm$1.9 \\
V & IV & IVT & 37.7$\pm$3.0 & 49.1$\pm$1.6 \\
V & VT & IVT & 37.7$\pm$1.9 & 48.8$\pm$2.3 \\
T & VT & IVT & 38.8$\pm$2.4 & 49.2$\pm$2.1 \\
\textbf{T*} & \textbf{IT*} & \textbf{IVT*} & \textbf{38.8$\pm$2.8} & \textbf{49.7$\pm$1.9} \\
\bottomrule
\end{tabular}%
}
\end{table}
\begin{table}[!htbp]
\centering
\small
\renewcommand{\arraystretch}{1.15}
\caption{Sensitivity of CurConMix performance to mixup parameters on CholecT45 and LLS48}
\label{tab:mixup_sensitivity_matrix}
\resizebox{\columnwidth}{!}{%
\begin{tabular}{@{}lcccccccc@{}}
\toprule
& \multicolumn{4}{c}{\textbf{Feature mixup } \(\alpha_{\text{feat}}\) \,(\(\alpha_{\text{input}}=0.4\))} 
& \multicolumn{4}{c}{\textbf{Input mixup } \(\alpha_{\text{input}}\) \,(\(\alpha_{\text{feat}}=0.4\))} \\
\cmidrule(lr){2-5}\cmidrule(lr){6-9}
\textbf{Dataset} & 0.1 & 0.2 & 0.3 & \textbf{0.4}$^{*}$ & 0.1 & 0.2 & 0.3 & \textbf{0.4}$^{*}$ \\
\midrule
CholecT45 & 38.6 & 38.0 & 38.0 & \textbf{38.8} & 37.9 & 37.9 & 38.1 & \textbf{38.8} \\
LLS48     & 49.5 & 49.2 & 48.6 & \textbf{49.7} & 48.5 & 48.8 & 49.0 & \textbf{49.7} \\
\bottomrule
\end{tabular}%
}
\end{table}
\begin{table}[!htbp]
\centering
\small
\renewcommand{\arraystretch}{1.12}
\caption{
Effect of temporal pathway configurations on CholecT45 and LLS48
} 
\label{tab:temporal_stride_ablation_final}
\resizebox{\columnwidth}{!}{%
\begin{tabular}{c c c c c}
\toprule
\textbf{\(\gamma_k\)} & \textbf{\(\beta\)} & \textbf{Temporal Pathways (\(k\))} &
\textbf{CholecT45 (AP\textsubscript{\textit{IVT}})} &
\textbf{LLS48 (AP\textsubscript{\textit{IVT}})} \\
\midrule
\multirow{7}{*}{\checkmark} & \multirow{7}{*}{\checkmark}
& \{2,3,4\}     & 40.27$\pm$2.2 & 52.08$\pm$3.0 \\
& & \{4,5,6\}     & 41.04$\pm$2.7 & \textbf{56.57$\pm$3.0} \\
& & \{4,5,6,7\}   & \textbf{41.57$\pm$2.8} & 55.08$\pm$2.6 \\
& & \{4,6,9\}     & 41.29$\pm$1.7 & 52.35$\pm$2.2 \\
& & \{7,8,9,10\}     & 41.18$\pm$3.1 & 54.36$\pm$3.1 \\
& & \{2-10\}      & 40.80$\pm$2.5 & 54.36$\pm$3.1 \\
\bottomrule
\end{tabular}%
}
\end{table}
\begin{table*}[!t]
\centering
\caption{
Transfer learning from action to phase/step/task prediction on Cholec80 and LLS48
}
\label{tab:transfer_results_table}
\small
\setlength{\tabcolsep}{4.3pt}
\renewcommand{\arraystretch}{1.05}
\begin{tabular}{ll ccc|ccc}
\toprule
\multicolumn{2}{c}{} & \multicolumn{3}{c}{\textbf{Supervised Baselines}} & \multicolumn{3}{c}{\textbf{CurConMix (Ours)}} \\
\cmidrule(lr){3-5} \cmidrule(lr){6-8}
\textbf{Dataset} & \textbf{Metric} & \textbf{TeCNO} & \textbf{Not-End} & \textbf{Trans-SVNet} & \textbf{TeCNO} & \textbf{Not-End} & \textbf{Trans-SVNet} \\
\midrule
\multirow{4}{*}{Cholec80–\textit{Phase}} 
& Accuracy  & 91.33$\pm$0.51 & 91.68$\pm$0.40 & 89.67$\pm$0.91 & \textbf{93.33$\pm$0.53} & 93.27$\pm$0.59 & 93.32$\pm$0.26 \\
& Recall    & 90.58$\pm$1.23 & 89.95$\pm$1.14 & 88.93$\pm$1.97 & 93.20$\pm$0.52 & 92.65$\pm$0.90 & \textbf{93.31$\pm$0.56} \\
& Precision & 89.17$\pm$0.20 & 89.85$\pm$0.64 & 84.79$\pm$1.21 & 90.40$\pm$0.56 & \textbf{90.72$\pm$0.40} & 89.21$\pm$0.30 \\
& Jaccard   & 79.91$\pm$1.26 & 80.05$\pm$0.70 & 74.23$\pm$2.49 & \textbf{83.89$\pm$0.52} & 83.41$\pm$1.15 & 82.74$\pm$0.52 \\
\midrule
\multirow{4}{*}{LLS48–\textit{Step}} 
& Accuracy  & 85.57$\pm$1.77 & 85.30$\pm$1.28 & 82.53$\pm$1.07 & \textbf{85.80$\pm$0.55} & 85.33$\pm$1.10 & 84.19$\pm$1.11 \\
& Recall    & 82.86$\pm$2.87 & 82.60$\pm$1.67 & 78.36$\pm$1.69 & \textbf{85.04$\pm$2.15} & 84.01$\pm$1.72 & 81.85$\pm$3.06 \\
& Precision & 83.02$\pm$1.35 & 82.32$\pm$1.35 & 78.46$\pm$0.79 & 83.32$\pm$1.39 & \textbf{83.90$\pm$1.06} & 82.17$\pm$0.77 \\
& Jaccard   & 70.59$\pm$3.51 & 69.81$\pm$2.76 & 64.15$\pm$2.06 & \textbf{71.87$\pm$2.09} & 71.51$\pm$2.13 & 68.27$\pm$1.75 \\
\midrule
\multirow{4}{*}{LLS48–\textit{Task}} 
& Accuracy  & 67.38$\pm$2.15 & 68.91$\pm$2.22 & 66.07$\pm$2.65 & 64.30$\pm$3.12 & 68.55$\pm$2.36 & \textbf{69.05$\pm$2.27} \\
& Recall    & 57.19$\pm$3.73 & 57.40$\pm$3.03 & 55.16$\pm$3.99 & 50.80$\pm$3.73 & 56.97$\pm$4.48 & \textbf{59.81$\pm$4.47} \\
& Precision & 54.05$\pm$1.80 & 59.43$\pm$2.73 & 55.31$\pm$3.68 & 51.32$\pm$4.10 & \textbf{61.34$\pm$4.18} & 59.92$\pm$2.50 \\
& Jaccard   & 38.45$\pm$2.36 & 40.62$\pm$2.30 & 37.94$\pm$2.41 & 33.48$\pm$3.14 & 40.47$\pm$4.18 & \textbf{42.31$\pm$2.81} \\
\bottomrule
\end{tabular}
\end{table*}
\subsection{Hyperparameter Robustness and Temporal Sensitivity}
\label{sec:hyperparameter_sensitivity}
We evaluated the robustness of the \textbf{CurConMix} and its temporal extension \textbf{CurConMix+} to variations in curriculum sequence, mixup parameters, and temporal pathway configurations. 
This analysis ensures that the observed improvements stem from principled design choices rather than dataset-specific or hyperparameter-dependent artifacts.

\textbf{Curriculum sequence.}
Table~\ref{tab:curriculum_order_mean_only} compares stage orders on CholecT45 and LLS48. The proposed progression—$\langle$target$\rangle \rightarrow \langle$instrument, target$\rangle \rightarrow \langle$instrument, verb, target$\rangle$—achieves the best AP\textsubscript{IVT} (38.8\% and 49.7\%). In contrast, curricula that start from \textit{verb} perform worst, consistent with surgical reasoning that a verb is defined by instrument--target interaction and is ambiguous without both.

\textbf{Mixup parameters.}
Table~\ref{tab:mixup_sensitivity_matrix} presents the effect of mixup coefficients $\alpha_{\text{feat}}$ and $\alpha_{\text{input}}$ within $[0.1, 0.4]$. The key finding is that performance remains highly stable, with variations below 0.5 percentage points across both datasets, indicating that CurConMix is robust to mixup intensity. Such robustness is significant. It reduces the need for dataset-specific calibration and demonstrates that the framework's improvements arise from its core design rather than from hyperparameter tuning. The baseline configuration ($\alpha_{\text{feat}} = \alpha_{\text{input}} = 0.4$), which achieves the highest accuracy on CholecT45 (38.8\%) and LLS48 (49.7\%), was selected from this stable range.

\textbf{Temporal pathway sensitivity.} Table~\ref{tab:temporal_stride_ablation_final} summarizes the effect of temporal pathway configurations in \textbf{CurConMix+} across CholecT45 and LLS48. The best performance is obtained with mid-range temporal windows, with the configuration \{4,5,6,7\} achieving 41.6\% AP\textsubscript{\textit{IVT}} on CholecT45 and \{4,5,6\} achieving 56.6\% on LLS48. Short-term pathways ({2,3,4}) capture only local motion cues, limiting temporal coherence, while excessively long spans ({7–10}) blur action boundaries due to rapid context shifts and overlapping gestures. Therefore, the mid-term windows provide the optimal balance between temporal precision and contextual continuity. The consistency of this trend across both datasets suggests that this mid-range configuration is a robust and generalizable choice, rather than a dataset-specific artifact.

\subsection{Transfer Learning from Action Triplets to High-Level Surgical Understanding} 
To assess whether fine-grained representations generalize to higher workflow levels, we conducted a transfer learning experiment. We first used an encoder trained within the \textbf{CurConMix} framework on surgical action triplet recognition. 
Then, to strictly evaluate the intrinsic quality of these learned features, we froze the trained backbone and trained only the temporal models (TeCNO, Not-End, and Trans-SVNet) on higher-level recognition tasks, including \textit{phase} recognition on Cholec80 and \textit{step} and \textit{task} recognition on LLS48. 
This setup examines whether the action-centric features can serve as a robust, fixed foundation for high-level procedural understanding without any task-specific fine-tuning of the spatial encoder.

The results, summarized in Table~\ref{tab:transfer_results_table}, show that CurConMix pretraining provides clear and consistent benefits for Cholec80-Phase and LLS48-Step. On these tasks, the CurConMix-pretrained models outperform the supervised baselines across all metrics and temporal models. For instance, on Cholec80, CurConMix with TeCNO achieves 93.33\% accuracy, significantly exceeding the supervised model (+2.00\% Acc, +3.98\% Jaccard).
On the more complex LLS48-Task, the results highlight that CurConMix's effectiveness is influenced by the temporal model. Here, the CurConMix-pretrained Trans-SVNet achieves the highest overall performance (69.05\% Acc, 42.31\% Jaccard), surpassing the best supervised baseline (68.91\% Acc, 40.62\% from Not-End).

These results indicate that representations trained on action-level semantics capture transferable compositional dependencies between instruments, verbs, and targets. Overall, these findings demonstrate that action-level pretraining via \textbf{CurConMix} provides a strong foundation for hierarchical surgical understanding, bridging fine-grained motion reasoning with coarse-grained procedural representation.

\section{Conclusion}
This study addressed three central challenges in surgical action triplet recognition---class imbalance, subtle visual ambiguity, and complex semantic dependencies among components.
We introduced \textbf{CurConMix}, a spatial representation framework that learns discriminative and semantically structured features. Its core, curriculum-guided contrastive learning, is made more effective through progressive hard-pair sampling and feature-level mixup.
Its temporal extension, \textbf{CurConMix+}, integrates a Multi-Resolution Temporal Transformer (MRTT) that adaptively fuses motion cues across time, unifying spatial–semantic learning with temporal reasoning for coherent triplet recognition.
Comprehensive evaluations on two distinct surgical domains, CholecT45 (cholecystectomy) and LLS48 (hepatectomy), demonstrate that \textbf{CurConMix+} achieves superior performance compared with existing methods. Qualitative visualizations further confirmed that the model learned to focus on clinically relevant tool–tissue interactions, validating the quality of spatial–semantic learning. Moreover, our framework exhibits low sensitivity to hyperparameter variations, enabling strong generalization across datasets.
The robustness and intrinsic quality of these representations were further validated through cross-level transfer experiments. A frozen \textbf{CurConMix} backbone—without any task-specific fine-tuning—consistently outperformed supervised baselines in \textit{phase}, \textit{step}, and \textit{task} recognition. These results also confirm that fine-grained \textit{action}-level understanding provides a transferable and powerful foundation for higher-level procedural reasoning.
Finally, we introduced the \textbf{LLS48} dataset, a hierarchically annotated benchmark for laparoscopic left lateral sectionectomy aligned with the SAGES taxonomy. This dataset not only expands the scope of surgical workflow analysis to complex hepatic procedures but also enables rigorous cross-level validation.
Together, \textbf{CurConMix}, \textbf{CurConMix+}, and \textbf{LLS48} establish a unified framework and benchmark for hierarchical surgical understanding, paving the way toward clinically grounded, generalizable models for real-world surgical AI.

\bibliographystyle{IEEEtran}
\bibliography{references}

\end{document}